\documentclass[preprint,3p,times,twocolumn]{elsarticle}

\usepackage{prletters}

\usepackage{geometry}
\usepackage{txfonts}
\usepackage{amssymb}
\usepackage{amsmath} 
\usepackage{soul}  
\usepackage{graphicx} 
\usepackage{sidecap}
\usepackage{booktabs}
\usepackage[breaklinks]{hyperref}
\usepackage{url}
\usepackage[labelfont=bf]{caption}
\usepackage{chngcntr}
\hypersetup{colorlinks = true,linkcolor = blue,anchorcolor =red,citecolor = blue,filecolor = red,urlcolor = red,pdfauthor=author}

\journal{Pattern Recognition Letters}

\begin{document}

\begin{frontmatter}

\title{A Too-Good-to-be-True Prior to Reduce Shortcut Reliance\tnoteref{t1, t2}}
\tnotetext[t1]{Preprint. Under consideration at Pattern Recognition Letters.}
\tnotetext[t2]{}

\author[hseuni]{Nikolay Dagaev\corref{corauth}}
\ead{ndagaev@hse.ru}
\cortext[corauth]{Corresponding author}
\address[hseuni]{School of Psychology, HSE University, Moscow, Russia}
\author[ucl]{Brett D.~Roads}
\author[ucl]{Xiaoliang Luo}
\author[ucl]{Daniel N.~Barry}
\author[inm,isn]{Kaustubh R.~Patil}
\author[ucl,ati]{Bradley C.~Love}
\address[ucl]{Department of Experimental Psychology, University College London, London, United Kingdom}
\address[inm]{Institute of Neuroscience and Medicine, Brain and Behaviour (INM-7), Research Center Jülich, Jülich, Germany}
\address[isn]{Institute of Systems Neuroscience, Medical Faculty, Heinrich Heine University Düsseldorf, Düsseldorf, Germany}
\address[ati]{The Alan Turing Institute, London, United Kingdom}


\begin{abstract}
Despite their impressive performance in object recognition and other tasks under standard testing conditions, deep networks often fail to generalize to out-of-distribution (o.o.d.) samples. One cause for this shortcoming is that modern architectures tend to rely on “shortcuts” – superficial features that correlate with categories without capturing deeper invariants that hold across contexts. Real-world concepts often possess a complex structure that can vary superficially across contexts, which can make the most intuitive and promising solutions in one context not generalize to others. One potential way to improve o.o.d. generalization is to assume simple solutions are unlikely to be valid across contexts and avoid them, which we refer to as the \emph{too-good-to-be-true prior}. A low-capacity network (LCN) with a shallow architecture should only be able to learn surface relationships, including shortcuts. We find that LCNs can serve as shortcut detectors. Furthermore, an LCN's predictions can be used in a two-stage approach to encourage a high-capacity network (HCN) to rely on deeper invariant features that should generalize broadly. In particular, items that the LCN can master are downweighted when training the HCN. Using a modified version of the CIFAR-10 dataset in which we introduced shortcuts, we found that the two-stage LCN-HCN approach reduced reliance on shortcuts and facilitated o.o.d. generalization.
\end{abstract}



\end{frontmatter}


\section{Introduction}
\label{sec:Intro}

\emph{``If you would only recognize that life is hard, things would be so much easier for you."---Louis D. Brandeis}
\newline

Deep convolutional neural networks (DCNNs) have achieved notable success in image recognition, sometimes achieving human-level performance or even surpassing it \cite{he2015delving}. However, DCNNs often suffer when out-of-distribution (o.o.d.) generalization is needed, that is, when training and test data are drawn from different distributions \cite{beery2018recognition, geirhos2019generalisation, geirhos2020shortcut}. This limitation has multiple consequences, such as susceptibility to adversarial interventions \cite{szegedy2013intriguing, goodfellow2014explaining, hendrycks2019natural} or to previously unseen types of noise \cite{geirhos2019generalisation, hendrycks2019benchmarking}.

Failure to generalize o.o.d. may reflect the tendency of modern network architectures to discover simple features, so-called ``shortcut'' features \cite{geirhos2020shortcut, shah2020pitfalls}. While the perils of overly-complex solutions are well appreciated, \textbf{overly-simplistic solutions should be viewed with equal skepticism}. In this work, we assume that features that are easy to learn are likely too good to be true. For instance, a green background may be highly correlated with the ``horse'' category, but green grass is not a central feature. A horse detector relying on such simplistic features--i.e. shortcuts--may perform well when applied in Spain---where the training set originates---but will fail when deployed in snow-covered Siberia. In effect, shortcuts are easily discovered by a network but may be inappropriate for classifying items in an independent set where superficial features are distributed differently than in the training set. Thus, the sensitivity to shortcuts may have far-reaching and dangerous consequences in applications, like when the pneumonia predictions of a system were based on a metal token placed in radiographs \cite{zech2018variable}.

In general, one cannot \emph{a priori} know whether shortcuts will be helpful or misleading, nor can shortcut learning be reduced to overfitting the training data. While overfitting can be estimated using an available test set from the same distribution, assessing shortcuts depends on all possible unseen data. A model relying on shortcuts can show remarkable human-level results on test sets where shortcut features are distributed identically to the training set (i.i.d.), but fail dramatically on o.o.d. test sets where shortcuts are missing or misleading \cite{recht2019imagenet}.

\begin{figure}[ht]
\centering
\includegraphics[width=\columnwidth]{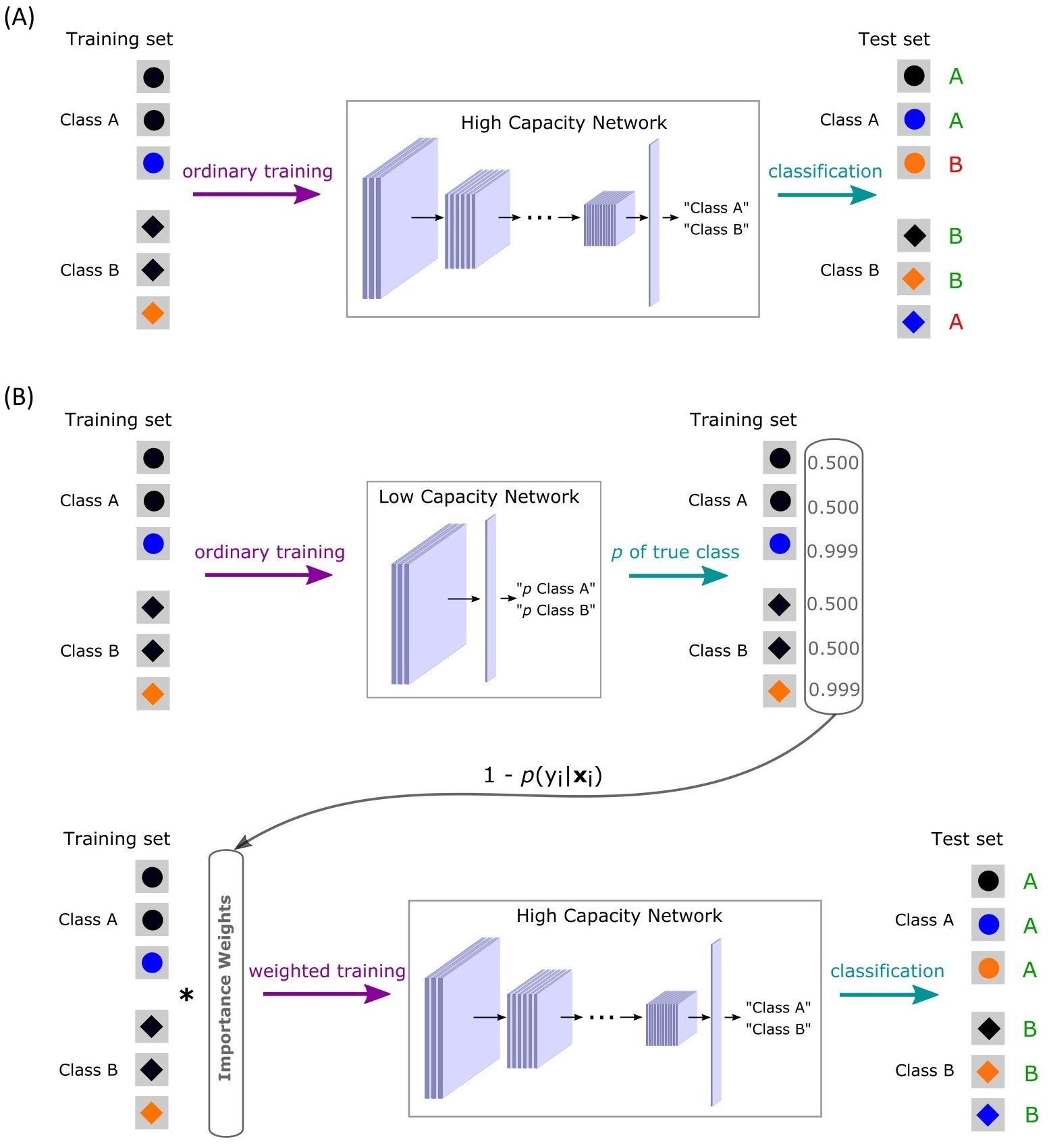}
\caption{The standard and too-good-to-be-true prior approaches to learning. (A) In the standard approach, a single high-capacity network (HCN) is trained and is susceptible to shortcuts, in this case relying on color as opposed to shape. Such a network will generalize well to i.i.d. test items but fail on o.o.d. test items (the last item for each class; shown in red). (B) In contrast, implementing the too-good-to-be true prior by pairing a low-capacity network (LCN) with an HCN leads to successful i.i.d. and o.o.d. generalization. Items that the LCN can master, which may contain shortcuts, are downweighted when the HCN is trained, which should reduce shortcut reliance and promote use of more complex and invariant features by the HCN.}
\label{fig:training-schemes}

\end{figure}

Shortcuts can adversely affect generalization even when they are not perfectly predictive. Because shortcuts are easily learned by DCNNs, they can be misleading even in the presence of more reliable but complex features \cite{hermann2020shapes}. To illustrate, shape may be perfectly predictive of class membership but networks may rely on color or other easily accessed features like texture \cite{geirhos2018imagenet, brendel2019approximating} when tested on novel cases (see Figure~\ref{fig:training-schemes}A).

Although what is and is not a shortcut cannot be known with perfect confidence, all shortcuts are simple. We find it unlikely that difficult learning problems will have trivial solutions when they have not been fully solved by brains with billions of neurons shaped by millions of years of natural selection nor by engineers working diligently for decades. Based on this observation, we are skeptical of very simple solutions to complex problems and believe they will have poor o.o.d. generalization. This inductive bias, which we refer to as the ``too-good-to-be-true prior'', can be incorporated into the training of DCNNs to reduce shortcut reliance and promote o.o.d. generalization. At its heart, the too-good-to-be true prior is a belief about the relationship between the world, models, and machine learning problems, which places limits on Occam's razor. Several recent contributions on o.o.d. generalization are consistent with the too-good-to-be-true prior \cite{clark2020learning, nam2020learning, sanh2020learning}. In various ways, these authors suggest that simple solutions should be treated with caution and avoided.

How does one identify solutions that are probably too-good-to-be-true? Here we suggest to make use of a learning system wittingly simplistic for the problem at hand and capable of only trivial solutions, which includes shortcuts. First, we show that such low-capacity systems can be used to detect shortcuts in a dataset. Second, we suggest a simple and general method aimed at discarding training examples that are suspected of containing shortcuts. We hypothesize that, in order to prevent shortcut learning by a high-capacity network (HCN), the predictions of a much simpler, low-capacity network (LCN) could be used to guide the training of the target network. Namely, a trained LCN would provide high-probability predictions precisely for the training items containing these shortcuts. Such probabilistic predictions can be transformed into importance weights (IWs) for training items, and these IWs can be further used in a loss function for training an HCN by \textbf{downweighting the shortcut items} (Figure~\ref{fig:training-schemes}B). We demonstrate our method’s efficiency by applying it to all possible CIFAR-10-based binary classification problems with synthetic shortcuts, permitting well-controlled experiments.

\section{Related Work}
\label{sec:RelatedWork}
\textbf{Shortcut learning and robust generalization.} Multiple approaches have been suggested for preventing shortcut reliance and increasing generalization robustness in deep neural networks \cite{geirhos2020shortcut}. We succinctly summarize eleven studies which we find most relevant to our work in Table~\ref{table:1}, using two criteria: whether an approach (1) assumes that simple solutions are probably shortcuts and (2) requires \emph{a priori} knowledge of the shortcut. We also specify the task domains considered.

\begin{table}[htbp]
\caption{Overview of approaches to preventing shortcut reliance most relevant to the present study. Task abbreviations: IR - image recognition, AR - action recognition, NLI - natural language inference, QA - question answering, VQA - visual question answering.}
\label{table:1}
\centering
\resizebox{\columnwidth}{!}{\begin{tabular}{l c c c} 
 \toprule
  \multirow[t]{2}{*}{\textbf{Approach}} & \shortstack{Simple solutions are shortcuts} & \shortstack{Requires knowing a shortcut} & \multirow[t]{2}{*}{Task} \\ %
 \midrule
 Reduce texture bias \cite{geirhos2018imagenet} & No & Yes & IR\\
 DRiFt \cite{he2019unlearn} & Yes & Yes & NLI \\ 
 Don’t take the easy\dots \cite{clark2019don} & Yes & Yes & QA, VQA, NLI \\
 REPAIR \cite{li2019repair} & No & No & IR, AR \\
 Learning not to learn \cite{kim2019learning} & No & Yes & IR \\
 RUBi \cite{cadene2019rubi} & No & Yes & VQA \\ 
 ReBias \cite{bahng2020learning} & No & Yes & IR, AR \\ 
 LfF \cite{nam2020learning} & Yes & No & IR \\
 DIBS \cite{sinha2020dibs} & No & No & IR \\
 Learning from others\dots \cite{sanh2020learning} & Yes & No & QA, NLI \\
 MCE \cite{clark2020learning} & Yes & No & IR, VQA, NLI \\
 \textbf{Our approach} & Yes & No & IR \\
 \bottomrule
\end{tabular}}
\end{table}

In contrast to \citet{nam2020learning}, we use an LCN, not a full-capacity target model, to identify shortcuts and train the LCN separately from the target HCN. In comparison to \citet{clark2020learning}, our approach is less demanding computationally and, again, the LCN and HCN are trained separately. We demonstrate that, in our particular implementation of the too-good-to-be-true prior, the limited capacity of a secondary model plays a key role, thus complementing the results of \citet{sanh2020learning}. We also extend the findings of Sanh et al., who introduced a similar de-biasing approach in the language domain, to the domain of image recognition. In contrast to all of the aforementioned studies, we show that an LCN can be employed to detect the presence of shortcuts in a dataset. Further, we empirically examine the relationship between the difficulty of a classification problem and the effectiveness of shortcut-avoiding training via our two-stage LCN-HCN procedure.

\citet{huangRSC2020} suggested a heuristic, Representation Self-Challenging (RSC), to improve o.o.d. generalization in image recognition. This method impedes predicting class from features most correlated with it and thus encourages a DCNN to rely on more complex combinations of features. RSC, however, is not directly designed to prevent shortcut learning but rather attempts to expand the set of features learned.

\textbf{Sample weighting.} Re-weighting of data samples is a well-known approach to guiding the training of DCNNs and machine learning models in general, and corresponding methods differ in terms of which examples must be downweighted/emphasized. Some authors suggested to mitigate the impact of easy examples and focus on hard ones \cite{malisiewicz2011ensemble, shrivastava2016training}. In contrast, in other research directions, such as curriculum learning \cite{bengio2009curriculum, hacohen2019power, wu2020curricula} and self-paced learning \cite{kumar2010self, meng2015objective}, it is recommended to stress easy examples early in training. It was also shown that the self-paced and curriculum learning can be combined \cite{jiang2015self}.

Although in our two-stage LCN-HCN procedure we assign weights to the training items, this method is fundamentally different from typical re-weighting schemes. Stemming from the too-good-to-be-true prior, our approach exploits not the predictions of the target network itself but of an independent simpler network (LCN). In other words, we are not interested in the difficulty of an item per se, but in whether this item can be mastered through simple means.

\section{Example applications of the too-good-to-be-true prior}
\label{sec:ExamplesApps}

Below, in the context of image recognition tasks, we illustrate the too-good-to-be-true prior with two example applications: (1) detecting the presence of a shortcut in a dataset and (2) training a de-biased model. Both examples rely on an LCN being limited to learning a superficial shortcut as opposed to a deeper invariant representation.

\subsection{The performance of a low-capacity network as an early warning signal}
\label{subsec:EarlyWarningSignal}

Considering that an LCN is only able to discover simple, probably shortcut, solutions, its high performance on a dataset may indicate the presence of a shortcut. When an LCN achieves a performance level comparable to an HCN, this should serve as a warning signal that the HCN may have succumbed to a shortcut \cite{hermann2020shapes}. In such cases, the HCN will likely fail to generalize robustly.

Here we present an illustrative example of such an application of the too-good-to-be-true prior: We trained an LCN (softmax regression) and an HCN (ResNet-56; \cite{he2016deep}) to classify the colored MNIST dataset. The latter was implemented exactly as in \cite{li2019repair}, with the standard deviation of color set to 0.1. Both networks were trained with stochastic gradient descent for 50 epochs (learning rate was set to 0.1 and mini-batch size was set to 256).

The presence of a shortcut indeed impairs o.o.d. generalization: both the LCN and HCN, after being trained on the colored data (shortcut is present), reveal poor performance on the regular test data (no shortcut). The poor o.o.d. generalization is particularly pronounced for the HCN - the difference between colored (0.999) and regular (0.384) test accuracies is 0.615. The difference in performance between the LCN and HCN after training and testing on the colored data is only 0.029, which serves as a warning that there may be a shortcut present. 

We also observed the same pattern for a stylized version of Tiny ImageNet \cite{wu2017tiny}, where we introduced a texture shortcut: the LCN and HCN showed relatively close high accuracies on stylized data (0.701 and 0.880, respectively) while performing dramatically different on regular data (0.085 and 0.414, respectively). The complete results for both datasets, as well as training details and architectures used for Tiny ImageNet, are in \ref{appendix:A}.

\subsection{Utilizing predictions of a low-capacity network to navigate the training of a high-capacity network}
\label{subsec:IWsforHCN}

Next, we demonstrate that it is possible to make use of an LCN to avoid learning the shortcut by an HCN. Reliable features necessary for a robust generalization are relatively high-level and shortcuts are usually low-level characteristics of an image. Given this assumption, the LCN will primarily produce accurate and confident predictions for images containing shortcuts.

Given a training dataset $\mathcal{D} = \{\mathbf{x}_{i}, y_{i}\}$, the corresponding IW ($w_i$) for a training image $\mathbf{x}_i$ is its probability of misclassification as given by an LCN,
\begin{equation}
    w_{i} = 1 - p(y_i|\mathbf{x}_i).
\end{equation}

IWs are then employed while training an HCN: for every training image, the corresponding loss term is multiplied by the IW of this image. We normalize IWs with respect to a mini-batch: IWs of samples from a mini-batch are divided by the sum of all IWs in that mini-batch. The mini-batch training loss $L_{\mathcal{B}}$ is thus the following:
\begin{equation}
    L_{\mathcal{B}} = \sum_{k \in \mathcal{B}} \widetilde{w}_{k}L_{k}, 
\end{equation}
where $L_{k}$ indicates the loss of the $k$th sample in the mini-batch. The mini-match normalized IW is 
\begin{equation}
    \widetilde{w}_{j} = \frac{w_{j}}{\sum_{k \in \mathcal{B}} w_{k}}.
\end{equation}

\subsubsection{Overview of experiments}
\label{subsubsec:ExpsOverview}
Generally, whether a dataset contains shortcuts is not known beforehand. In order to overcome this issue and test the too-good-to-be-true prior, we introduced synthetic shortcuts into a well-known dataset \cite[cf.][]{MALHOTRA202057}. We then applied our approach and investigated whether it was able to avoid reliance on these shortcuts and generalize o.o.d. This testing strategy allowed us to run well-controlled experiments and quantify the effects of our method.

We ran a set of experiments on all possible pairs of classes from the CIFAR-10 dataset \cite{krizhevsky2009learning}. In every binary classification problem, a synthetic shortcut was introduced in each of the two classes. To have a better understanding of our method’s generalizability, we investigated two opposite types of shortcuts as well as two HCN architectures, ResNet-56 \cite{he2016deep} and VGG-11 \cite{simonyan2014very}. Note that our too-good-to-be-true prior is readily applicable to multi-class problems.

For both shortcut types and both HCN architectures, we expected the two-stage LCN-HCN procedure to downweight the majority of shortcut images. Therefore, compared to the ordinary training procedure, better performance should be observed when shortcuts in a test set are misleading (i.e., o.o.d. test set). We also expected that the two-stage LCN-HCN procedure may suppress some non-shortcut images. Thus, a slightly worse performance was expected for a test set without shortcuts as well as for a test set with helpful shortcuts (i.e., i.i.d. test set).

The main objective of these experiments was to compare an ordinary and a weighted training procedure in terms of the susceptibility of resulting models to the shortcuts. However, crucially for our idea of the too-good-to-be-true prior, it was also important to validate our reasoning concerning the key role of the low-capacity network in the derivation of useful IWs. For this purpose, we introduced another training condition where IWs were obtained from probabilistic predictions of the same HCN architecture as the target network. We refer to the IWs obtained from an HCN as HCN-IWs and to the IWs obtained from an LCN as LCN-IWs. We expected HCN-IWs either to fail to suppress shortcut images, resulting in poor performance on a test set with misleading shortcuts (o.o.d. test set), or to equally suppress both shortcut and non-shortcut images, resulting in poor performance on any test data. Using HCN-IWs mirrors approaches that place greater emphasis on challenging items.

\subsubsection{Shortcuts}
\label{subsubsec:Shortcuts}
For the sake of generality, we introduced two shortcut types: the ``local'' was salient and localized, and the ``global'' was subtle and diffuse. The local shortcut was intended to capture real-world cases such as a marker in the corner of a radiograph \cite{zech2018variable} and the global was intended to capture such situations as subtle distortions in the lens of a camera.

\begin{figure}[htbp]
\centering
\scalebox{.7}{\includegraphics[width=\columnwidth]{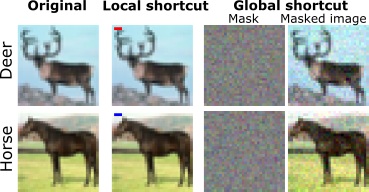}} %
\caption{Examples of shortcut types used in our experiments: local and global. The global shortcut is subtle to humans, so original images and additive masks are also depicted. For the subset of images containing a shortcut, a network could learn to rely on these superficial features at the expense of more invariant properties, which has consequences for generalization.}
\label{fig:example-shortcuts}
\vskip -0.1in
\end{figure}

The local shortcut was a horizontal line of three pixels, red for one class and blue for the other (Figure ~\ref{fig:example-shortcuts}, second column). The location of the line was the same for all images: upper left corner. The shortcut was present in randomly chosen 30\% of training as well as validation images in each class.

The global shortcut was a mask of Gaussian noise, one per class (Figure~\ref{fig:example-shortcuts}, right). The mask was sampled from a multivariate normal distribution with zero mean and isotropic covariance matrix, with variance set to $25 \times 10^{-4}$, and then added to randomly chosen 30\% of training and validation images of a corresponding class. 

\begin{figure}[htbp]
\centering
\includegraphics[width=\columnwidth]{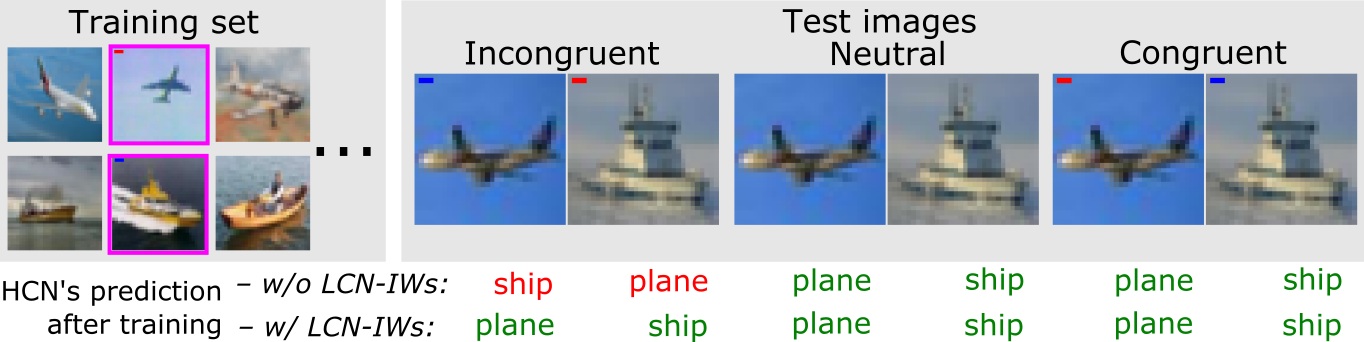}
\caption{Expected effects of the LCN-IWs training on classifying different test cases by the HCN. Correct HCN decisions are in green, incorrect are in red. Training images containing local shortcuts are outlined in magenta. Whereas ordinary (w/o LCN-IWs) training should lead to poor o.o.d. performance on Incongruent test items where the shortcut is now misleading, LCN-IWs should selectively downweight training items with the shortcut allowing the HCN to generalize well across the spectrum.}
\label{fig:experiment-predictions}
\vskip -0.1in
\end{figure}

Based on CIFAR-10 test images of the selected classes, we prepared three test sets for each shortcut type (examples shown in Figure~\ref{fig:experiment-predictions}). \emph{Congruent} (i.i.d.): all images contained shortcuts, each associated with the same class as in the training set. \emph{Incongruent} (o.o.d.): all images contained shortcuts but each shortcut was placed in the images of the opposite class compared to the training set. \emph{Neutral}: original CIFAR-10 images without shortcuts.

\subsubsection{Model architectures and training details}
\label{subsubsec:ArchsTrainDetails}
The LCN was a single convolutional layer of 4 channels with 3-by-3 kernels, a linear activation function, and without downsampling, followed by a fully-connected classification layer.

In two separate sets of simulations, we tested two different HCN architectures: ResNet-56 for CIFAR-10 \cite{he2016deep} and VGG-11 \cite{simonyan2014very}. The first two fully-connected layers of VGG-11 had 1024 units each and no dropout was used.

Network weights were initialized according to \cite{glorot2010understanding}. We used stochastic gradient descent to train both LCN and HCN. The initial learning rate was set to 0.01 for the LCN. The HCN’s initial learning rate was set to 0.01 for VGG-11 \cite{simonyan2014very} and to 0.1 for ResNet-56 \cite{he2016deep}. The HCNs were trained with a momentum of 0.9 and a weight decay of 5×10\textsuperscript{-4} for 150 epochs. To avoid overfitting, the HCN's performance on validation data (see below) was tested at each epoch, and best-performing parameters were chosen as the result of training. The LCN was trained for 40 epochs. For both LCN and HCN, the learning rate was decreased by a factor of 10 on epochs corresponding to 50\% and 75\% of the total duration of the training. Mini-batch size for both networks was set to 256.

For each class, the original 5,000 images from the CIFAR-10 training set were divided into 4,500 training images and 500 validation images. Thus, the training set of every class pair included 9,000 images and the validation set included 1,000 images.

IWs were introduced to the training process as described in the beginning of this section, and for every mini-batch a weighted-average loss was calculated. During ordinary training without IWs, a simple average loss was calculated.

For every binary classification problem, the results reported below are the averages of 10 independent runs.

\subsubsection{Results}
\label{subsubsec:Results}
The overall pattern of results was in accord with our predictions--downweighting training items that could be mastered by a low-capacity network reduced shortcut reliance in a high-capacity network, which improved o.o.d. generalization at a small cost in i.i.d. generalization. As no disagreement in the results of the two HCN architectures was observed, hereinafter we focus on ResNet-56, whereas results for VGG-11 can be found in \ref{appendix:C}. 

To compare the potentials of LCN-IWs and HCN-IWs to separate out shortcut-containing images, we used logistic regression to classify IWs as shortcut or regular. With the local shortcuts, classification accuracy averaged across 45 class-pairs was 91.53\% for LCN-IWs and 70.02\% for HCN-IWs, with corresponding 95\% CIs of [82.60, 93.45] and [70.01, 70.03]. With the global shortcuts, average classification accuracy was 90.4\% for LCN-IWs and 70.01\% for HCN-IWs, with corresponding 95\% CIs of [86.16, 94.66] and [70.00, 70.03]. Thus, the distribution of LCN-IWs much better discriminates between shortcut and regular images than the distribution of HCN-IWs. An example of HCN-/LCN-IWs distributions for a specific pair of classes can be found in \ref{appendix:B}.

\begin{figure*}[htbp]
\centering
\includegraphics[width=\textwidth]{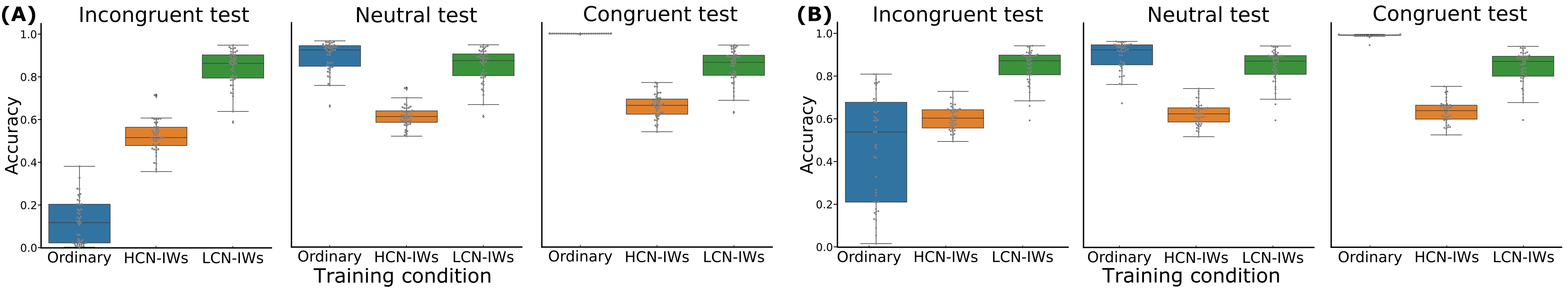}
\caption{Accuracies on incongruent, neutral, and congruent test sets after ordinary and HCN-/LCN-weighted training, with (A) local or (B) global shortcuts in training set. HCN is ResNet-56. Across shortcut types, LCN-IWs result in almost equally high accuracy on all sets. HCN-IWs constantly result in accuracies inferior to LCN-IWs; moreover, on neutral and congruent test sets, accuracies after HCN-IWs training are lower than after ordinary training. Thus, LCN-IWs are successful in avoiding shortcut reliance and preserving useful features, while HCN-IWs are not.}
\label{fig:test_accuracies}
\vskip -0.1in
\end{figure*}

Effects of the training condition (ordinary, HCN-IWs, and LCN-IWs) on the HCN's performance on every test set (incongruent, neutral, and congruent) are shown in Figure~\ref{fig:test_accuracies}. HCNs are prone to rely on our shortcuts, as evidenced by low incongruent accuracies and very high congruent accuracies after the ordinary training. Incongruent accuracies are improved after the LCN-IWs training, compared to those after the ordinary training. Importantly, after LCN-IWs training, incongruent, neutral, and congruent accuracies are all similarly high. Together, these results suggest that LCN-IWs are successful in reducing shortcut reliance in the target network.

Although exceeding performance on the incongruent test set after the ordinary training condition, incongruent accuracies after the HCN-IWs training are substantially lower than after the LCN-IWs training. The neutral and congruent accuracies are lower than in both ordinary and LCN-IWs training conditions. At the same time, incongruent accuracies are still noticeably lower than neutral and congruent conditions. 
These results indicate that HCN-IWs are not effective as LCN-IWs in suppressing shortcut learning.

\begin{figure*}[htbp]
\centering
\includegraphics[width=\textwidth]{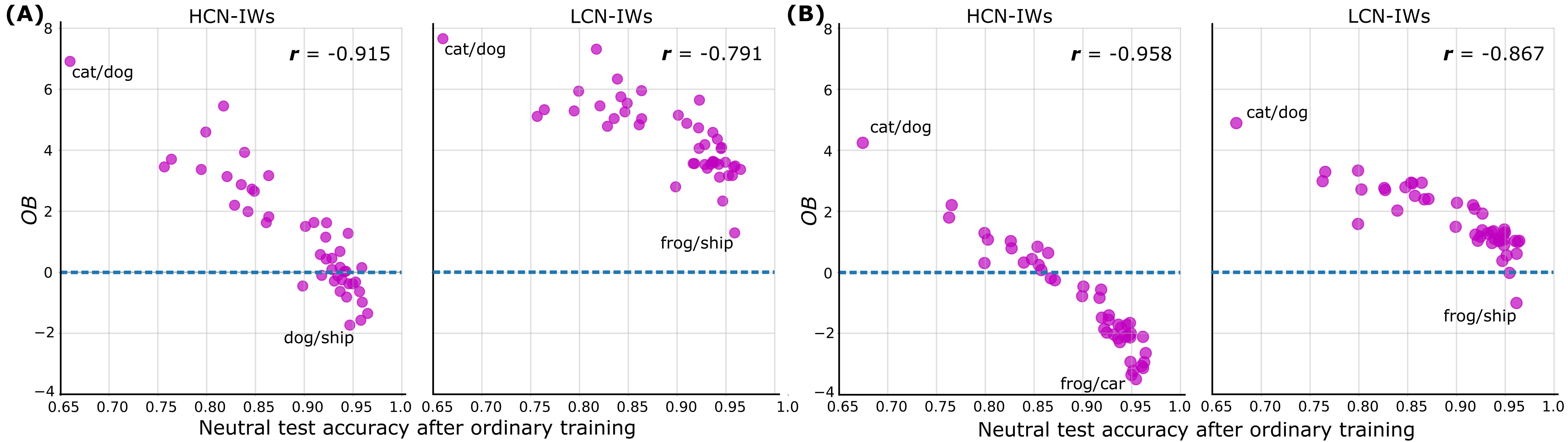}
\caption{Effects of the LCN-/HCN-IWs training procedure for each of the 45 class pairs depending on a difficulty of the respective binary classification problem. HCN is ResNet-56; (A) local and (B) global shortcut types are considered separately. The effects of training are represented by the Overall Benefit measure ($OB$; gain + loss; see Section \ref{subsubsec:Results}); the difficulty of a pair is represented by the neutral test accuracy after ordinary training. Recapitulating previous results, LCN-IWs are more effective than HCN-IWs. Furthermore, the easier learning problem, the less $OB$ from IWs: the relatively higher capacity of a network supplying IWs leads to downweighting non-shortcut items.}
\label{fig:overall_performance}
\vskip -0.2in
\end{figure*}

The main results shown in Figure~\ref{fig:test_accuracies} indicate that LCN-IWs reduce shortcut reliance with little cost to performance on other items, whereas HCN-IWs are less effective because they remove non-shortcut items as well. Key to the LCN-IW efficacy is properly matching network capacity to the learning problem. Out of the 45 classification pairs considered, there should be natural variation in problem difficulty that affects target network performance. In particular, we predict that overall benefit will be lower when the LCN performs better on a class pair, indicating that its capacity is sufficient to learn non-shortcut information for this classification task.

We define Overall Benefit ($OB$) as a combination of Gain ($G$) and Loss ($L$):
\begin{equation}
    OB = G + L,
\end{equation}
where
\begin{equation}
\begin{split}
    G =\;&\text{logit}(p(\text{correct} | \text{incongruent}, \text{IW}))\\ &- \text{logit}(p(\text{correct} | \text{incongruent}, \text{ordinary}))
\end{split}
\end{equation}
and
\begin{equation}
\begin{split}
    L =\;&\text{logit}(p(\text{correct} | \text{neutral}, \text{IW}))\\ &- \text{logit}(p(\text{correct} | \text{neutral}, \text{ordinary})).
\end{split}
\end{equation}
We compute average $OB$ for each class pair and contrast those against corresponding neutral test accuracies after ordinary training. The latter are introduced to reflect the default classification difficulty of each class pair. These comparisons are shown in Figure \ref{fig:overall_performance}. Two evident trends are important. First, recapitulating the previous results, LCN-IWs result in greater $OB$ than HCN-IWs. $OB$ corresponding to LCN-IWs is almost always positive, while $OB$ corresponding to HCN-IWs is often negative. Second, $OB$ is negatively correlated with the neutral test accuracy after ordinary training; that is, as the difficulty of a classification problem increases, benefits of using IWs generally increase as well. One possibility is that for easy to discriminate pairs, such as frog and ship, the LCN was able to learn non-shortcut information which reduced the overall benefit of the LCN-IWs.

\section{Discussion}
\label{section:Discus}

In general, using Occam's razor to favor simple solutions is a sensible policy. We certainly do not advocate for adding unnecessary complexity. However, for difficult problems that have evaded a solution, it is unlikely that a trivial solution exists. The problems of interest in machine learning have taken millions of years for nature to solve and have puzzled engineers for decades. It seems implausible that trivial solutions to such problems would exist and we should be skeptical when they appear.

For such difficult problems, we suggest adopting a \emph{too-good-to-be-true prior} that shies away from simple solutions. Simple solutions to complex problems are likely to rely on superficial features that are reliable within the particular training context, but are unlikely to capture the more subtle invariants central to a concept. To use a historic example, people had great hopes that the Perceptron \cite{rosenblattPerceptronProbabilisticModel1958}, a one-layer neural network, would master computer vision to only have their hopes dashed \cite{minskyPerceptronsIntroductionComputational1969}. When such simple systems appear successful, including on held-out test data, they are most likely relying on shortcuts that will not generalize out of sample on somewhat different test distributions, such as when a system is deployed.

We proposed and evaluated two simple applications of the too-good-to-be-true inductive bias. First, we made use of a low-capacity network (LCN) to detect the presence of a shortcut in a dataset. Second, we used an LCN to establish importance weights (IWs) to help train a high-capacity network (HCN). The idea was that the LCN would not have the capacity to learn subtle invariants but instead be reduced to relying on superficial shortcuts. For the second application, by downweighting the items that LCN could master, we found that the HCN was less susceptible to shortcuts and showed better o.o.d. generalization at little cost when misleading shortcuts were not present.

Although we evaluated the de-biasing application of the too-good-to-be-true prior on the CIFAR-10 dataset, the basic method of using an LCN to establish IWs for an HCN is broadly applicable. We considered two network architectures for the HCN, ResNet-56 and VGG-11, which both showed the same overall pattern of performance. Interestingly, ResNet-56 appeared more susceptible to shortcuts, perhaps because its architecture contains skip connections that are themselves a type of shortcut allowing lower-level information in the network to propagate forward absent intermediate processing stages.

One key challenge in our approach is matching the complexity of the LCN to the learning problem. When the LCN has too much capacity, it may learn more than shortcuts and downweight information useful to o.o.d. generalization (see Figure~\ref{fig:overall_performance}). It is for this reason that LCN-IWs are much more effective than HCN-IWs (see Figure~\ref{fig:test_accuracies}). Unfortunately, there is no simple procedure that guarantees selecting an appropriate LCN. The choice depends on one's beliefs about the structure of the world, the susceptibility of models to misleading shortcuts, and the nature of the learning problem. Nevertheless, reasonable decisions can be made. For example, we would be skeptical of a Perceptron that successfully classifies medical imagery, so it could serve as an LCN.

Since the too-good-to-be-true prior is a general inductive bias, our two-stage LCN-HCN approach is just one specific implementation of it and other techniques may be developed. The effectiveness of our two-stage approach should be evaluated in other tasks and domains outside computer vision. Further research should consider how to choose the architecture of an LCN and how the effectiveness of this architecture depends on different types of shortcuts. Finally, a promising direction is to use IWs to selectively suppress aspects of training items rather than downweighting entire examples. 

\section*{Acknowledgements}
\label{sec:Acknow}
This article is an output of a research project implemented as part of the Basic Research Program at the National Research University Higher School of Economics (HSE University). This work was supported by NIH Grant 1P01HD080679, Wellcome Trust Investigator Award WT106931MA, and Royal Society Wolfson Fellowship 183029 to B.C.L.

\bibliographystyle{elsarticle-harv} 
{\small\bibliography{prl_manuscript}}

\clearpage

\appendix
\counterwithin*{figure}{section}
\counterwithin*{table}{section}

\onecolumn

\section{Complete results for detecting the presence of shortcuts using the low-capacity network}
\label{appendix:A}
\subsection{Complete results for colored MNIST}
\label{subsec:ComplResCMNIST}

\begin{table*}[htbp]
\caption{LCN and HCN mean accuracies (averaged across 10 independent runs) on colored MNIST; standard deviations are in parentheses.}
\label{table:2}
\centering
\begin{tabular}{c c c c c}
 \toprule
 \textbf{Architecture} & \multicolumn{2}{c}{\textbf{Regular training}} & \multicolumn{2}{c}{\textbf{Colored training}} \\
 \cmidrule(r){2-5}
  & Regular test & Colored test & Colored test & Regular test \\ 
 \midrule
 LCN & 0.908 (0.012) & 0.899 (0.015) & 0.970 (0.007) & 0.617 (0.082) \\
 HCN & 0.996 (0.001) & 0.969 (0.017) & 0.999 (0.001) & 0.384 (0.128) \\ 
 \bottomrule
\end{tabular}
\end{table*}

\subsection{Training details and complete results for stylized Tiny ImageNet}
\label{subsec:StyTinyImNet}

We constructed a stylized version of Tiny ImageNet by following a generalization\footnote{\href{https://github.com/bethgelab/stylize-datasets}{https://github.com/bethgelab/stylize-datasets}} of the procedure described in \cite{geirhos2018imagenet}. Each of 200 classes was assigned its unique style and thus had a prominent texture shortcut.

The LCN was represented by a single 4-channel convolutional layer (3-by-3 kernels, linear activation function, no downsampling) followed by a fully-connected softmax classification layer. The HCN was represented by a 10-layer ResNet designed for Tiny ImageNet \cite{wu2017tiny}.
Both networks were trained for 40 epochs with a stochastic gradient descent, a momentum of 0.9, and a weight decay of 5×10\textsuperscript{-4}. A mini-batch size was set to 256. The initial learning rate was set to 0.001 and 0.1 for the LCN and HCN, respectively, and was decreased by a factor of 10 on epochs corresponding to 50\% and 75\% of the total duration of the network’s training.

\begin{table*}[htbp]
\caption{LCN and HCN mean accuracies (averaged across 10 independent runs) on stylized Tiny ImageNet; standard deviations are in parentheses.}
\label{table:3}
\centering
\begin{tabular}{c c c c c}
 \toprule
 \textbf{Architecture} & \multicolumn{2}{c}{\textbf{Regular training}} & \multicolumn{2}{c}{\textbf{Stylized training}} \\
 \cmidrule(r){2-5}
  & Regular test & Stylized test & Stylized test & Regular test \\
 \midrule
 LCN & 0.085 (0.002) & 0.013 (0.001) & 0.701 (0.003) & 0.007 (0.001) \\
 HCN & 0.414 (0.018) & 0.022 (0.003) & 0.880 (0.049) & 0.005 (0.001) \\ 
 \bottomrule
\end{tabular}
\end{table*}

\clearpage
\section{Results for VGG-11 as the high-capacity network}
\label{appendix:C}
\subsection{Separability of shortcut and regular items in LCN- and HCN-IWs}
\label{subsec:classify_IWs}
As described for ResNet-56 in the main text, we used logistic regression to classify IWs as shortcut or regular. When the shortcuts in the training set were local, classification accuracy averaged across 45 class-pairs was 91.53\% for LCN-IWs and 70.011\% for HCN-IWs, with corresponding 95\% CIs of [82.60, 93.45] and [70.01, 70.02]. When the shortcuts were global, average classification accuracy was 90.4\% for LCN-IWs and 70.011\% for HCN-IWs, with corresponding 95\% CIs of [86.16, 94.66] and [70.010, 70.024]. These results repeat the pattern we observed for ResNet-56, confirming that LCN-IWs, compared to HCN-IWs, much better distinguish shortcut images from regular ones. 

\subsection{Test accuracies and overall benefit}
\label{subsec:test_accs_ob}

\begin{figure*}[htbp]
\centering
\includegraphics[width=\textwidth]{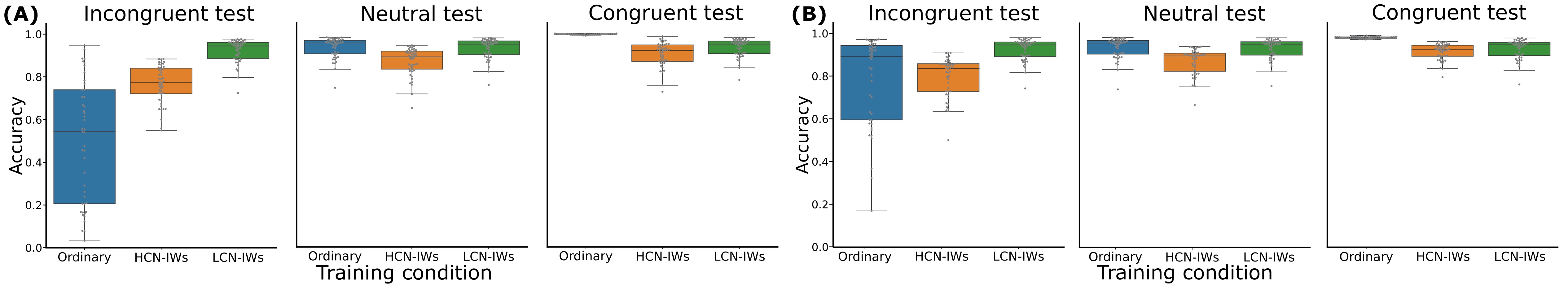}
\caption{Accuracies on incongruent, neutral, and congruent test sets after ordinary and HCN-/LCN-weighted training, with (A) local or (B) global shortcuts in training set. HCN is VGG-11. Across shortcut types, LCN-IWs result in almost identically high accuracy on all three test sets and thus, are successful in avoiding shortcut reliance. HCN-IWs constantly result in accuracies inferior to LCN-IWs; moreover, on neutral and congruent test sets, accuracies after HCN-weighted training are lower than after ordinary training. HCN-IWs, thus, are not as effective as LCN-IWs in avoiding shortcut reliance and also result in suppressing useful features. Together, these results indicate that the LCN-HCN two-stage approach is a valid representative of the too-good-to-be-true prior.}
\label{fig:test_accuracies_VGG-11}
\end{figure*}

\begin{figure*}[ht]
\centering
\includegraphics[width=\textwidth]{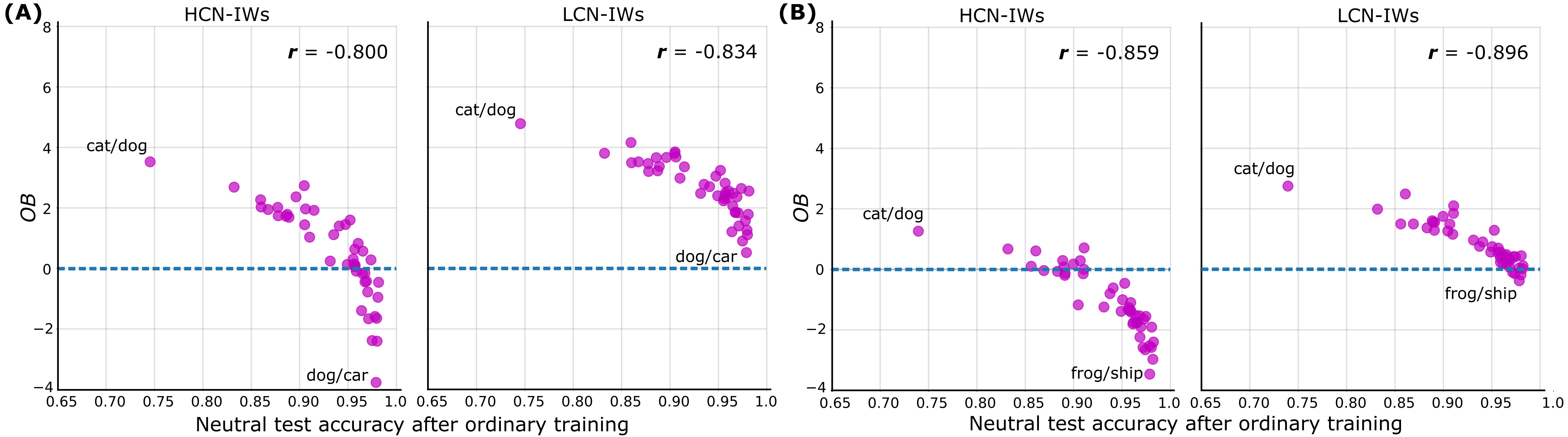}
\caption{Effects of the LCN-/HCN-IWs training procedure for each of the 45 class pairs depending on a difficulty of the respective binary classification problem. HCN is VGG-11; (A) local and (B) global shortcuts are considered separately. Effects of training are represented by the Overall Benefit measure ($OB$; gain + loss; see main text, Section \ref{subsubsec:Results}); the difficulty of a pair is represented by the neutral test accuracy after ordinary training on the training set. Recapitulating previous results, LCN-IWs are more effective than HCN-IWs. Furthermore, the easier learning problem, the less $OB$ from IWs because the relatively higher capacity of a producing IWs network leads to downweighting non-shortcut items.}
\label{fig:overall_performance_VGG11}
\end{figure*}

\clearpage
\section{Illustrative histograms of importance weights for a single pair of classes}
\label{appendix:B}

\begin{figure*}[htbp]
\centering
\includegraphics[width=\textwidth]{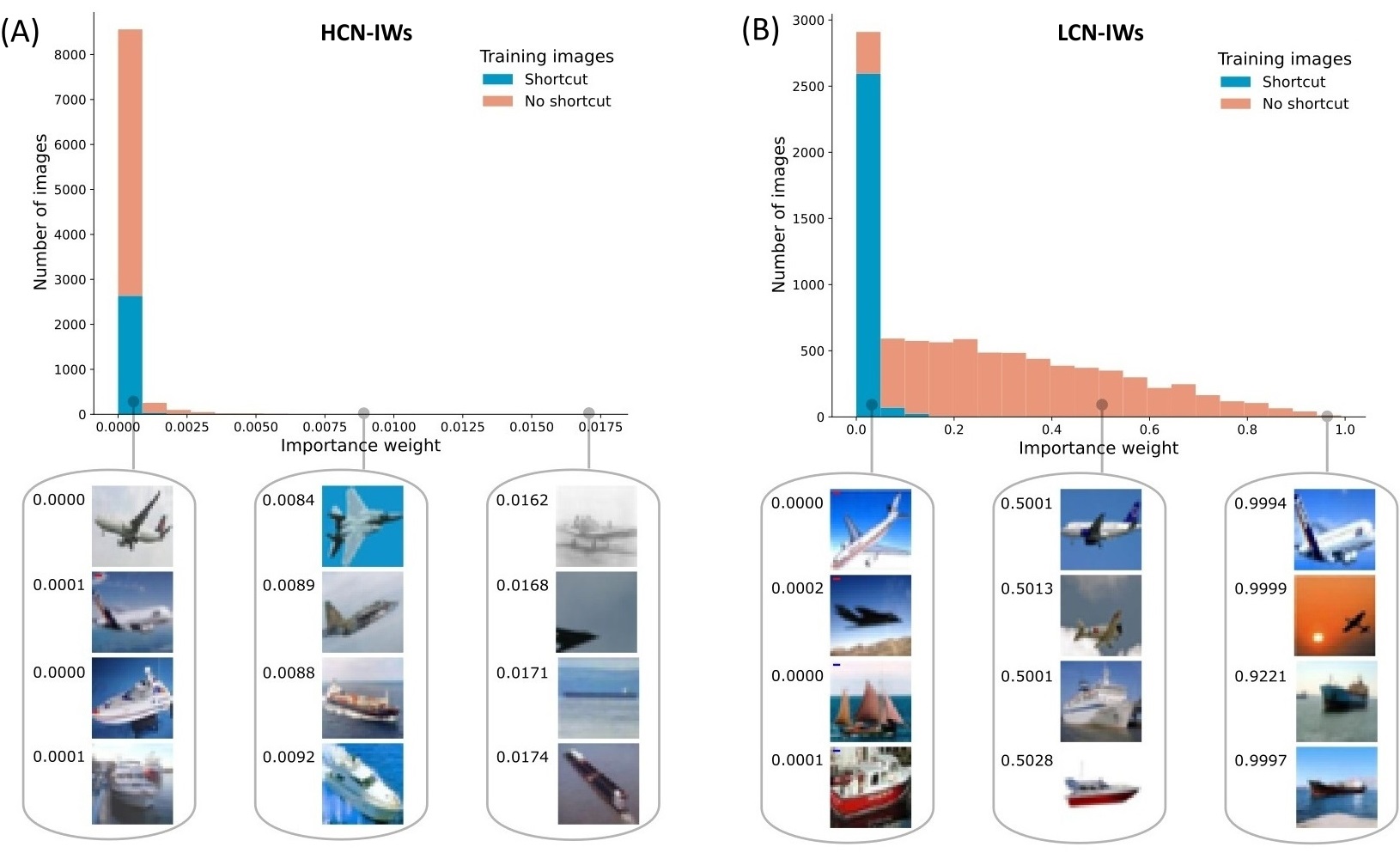}
\caption{Observed distributions of HCN-IWs (A) and LCN-IWs (B) for the plane/ship pair with the local shortcut in the training set. HCN is ResNet-56.
The near zero importance weights from the LCN were mostly for shortcut images.
As predicted, the LCN has the capacity to master images containing shortcuts but few other images, providing IWs for an HCN that reduce shortcut reliance, thereby implementing the too-good-to-be true prior.}
\label{fig:iw-distributions_examples}
\end{figure*}

\end{document}